%% file: main.tex
\documentclass[10pt,twocolumn,letterpaper]{article}

\usepackage[pagenumbers]{cvpr} 

\input{preamble}
\definecolor{cvprblue}{rgb}{0.21,0.49,0.74}
\usepackage[pagebackref,breaklinks,colorlinks,allcolors=cvprblue]{hyperref}
\usepackage{multirow}
\usepackage{graphicx}
\usepackage{subcaption} 
\usepackage{tikz}       
\usepackage{float}

\def\paperID{*****} 
\def\confName{CVPR}
\def\confYear{2026}

\title{Latent Domain Modeling Improves Robustness to Geographic Shifts}

\author{Ruth Crasto\\
Microsoft\\
{\tt\small ruthcrasto@microsoft.com}
\and
Esther Rolf\\
 University of Colorado Boulder\\
{\tt\small esther.rolf@colorado.edu}
}

\begin{document}
\maketitle
\input{sec/0_abstract}    
\input{sec/1_intro}
\input{sec/2_background}
\input{sec/3_method}
\input{sec/4_experiments}
\input{sec/5_discussion}
\input{sec/6_conclusion}
{
    \small
    \bibliographystyle{ieeenat_fullname}
    \bibliography{main}
}

\input{sec/X_suppl}

\end{document}

%% file: sec/0_abstract.tex
\begin{abstract}
Geographic distribution shift arises when the distribution of locations on Earth in a training dataset is different from what is seen at inference time. Using standard empirical risk minimization (ERM) in this setting can lead to uneven generalization across different spatially-determined groups of interest such as continents or biomes. The most common approaches to tackling geographic distribution shift apply domain adaptation methods using discrete group labels, ignoring geographic coordinates that are often available as metadata. On the other hand, modeling methods that integrate geographic coordinates have been shown to improve overall performance, but their impact on geographic domain generalization has not been studied. In this work, we propose a general modeling framework for improving robustness to geographic distribution shift. The key idea is to model continuous, latent domain assignment using location encoders and to condition the main task predictor on the jointly-trained latents. On four diverse geo-tagged image datasets with different group splits, we show that instances of our framework achieve significant improvements in worst-group performance compared to existing domain adaptation and location-aware modeling methods. In particular, we achieve new state-of-the-art results on two datasets from the WILDS benchmark. Our code is publicly available at:
\href{https://github.com/crastoru/wilds-geoshift}{https://github.com/crastoru/wilds-geoshift}.
\end{abstract}



%% file: sec/1_intro.tex
\section{Introduction}
\label{sec:intro}
Empirical risk minimization (ERM) assumes a set of independent and identically distributed (\textit{i.i.d}) training examples. Deep learning models trained using ERM are then expected to generalize to unseen data at inference time with the additional assumption that this data is drawn independently from the same distribution. In many real-world applications, these assumptions do not hold, resulting in degraded and unreliable model performance  \citep{dundar2007}.  

In this work, we focus on a particular way of violating these assumptions known as \textit{subpopulation shift} \citep{wilds}, where \textbf{1)} the training data is not \textit{i.i.d} but rather drawn a mixture of different distributions (domains), and \textbf{2)} the inference data is drawn from a different mixture of the same domains.

Subpopulation shift can arise naturally when training and inference data are drawn from different distributions of locations on Earth, which we refer to as \textit{geographic distribution shift}. When each data point is associated with a domain\footnote{In this paper, we use ``domain" and ``group" interchangeably.} that is correlated with its location---such as administrative region, biome, or rurality---a geographic distribution shift can induce a subpopulation shift over these domains. This shift often results in poor generalization within certain domains and presents an obstacle to global-scale model deployment, where the objective is typically to generalize well across all domains \citep{ekim2024ood}. For example, it has been observed that in critical applications such as crop yield estimation \cite{timl}, building damage detection \cite{robinson2022fast}, and biodiversity monitoring \cite{lange2023active}, training datasets in practice tend to be highly imbalanced across regions of interest, leading to degraded performance in under-represented regions.


A common solution to this problem is to apply standard domain adaptation or distributionally-robust optimization (DRO) techniques during training using discrete domain labels \citep{wilds, d3g}. However, these approaches fail to explicitly model the correlation between domain and location on Earth, and cannot capture intra-domain diversity or inter-domain similarities due to the discretization of the domain signal. Moreover, geographic coordinates that are commonly available as metadata are discarded.

On the other hand, there is an emerging body of work focused on modeling methods that explicitly incorporate geographic coordinates for improved learning of spatial context \cite{feifei2015, wrap, torchspatial}. These techniques rely on \textit{location encoders} to map geographic coordinates into a learning-friendly higher-dimensional latent space \citep{sphere2vec}. It has been shown that modeling spatial context using location encoders can lead to improvements in overall in-distribution and out-of-distribution performance on a variety of prediction tasks \citep{wrap, klemmer2023satclip, rao2025using, pm25} but may cause overfitting in some settings \cite{rao2025using, pm25}. While the impact of location-aware modeling on geographic domain generalization has not been studied, these findings suggest that such techniques may be beneficial in the subpopulation shift setting when applied in a principled way. 





\textbf{The central idea of this paper is to use location encoders to model continuous, latent domain assignment in place of discrete domain labels for improving robustness to geographically-induced subpopulation shift.} We propose a general modeling framework that involves \textbf{1)} learning domain latents using a domain prediction loss applied to a location encoder, and \textbf{2)} conditioning the main task predictor on these jointly-trained latents. Unlike prior works in domain adaptation, our approach directly models the correlation between domain and location on Earth. Furthermore, since location encoders are relatively lightweight, learning these domain latents can be done efficiently.

Our framework can flexibly incorporate any location encoder architecture or conditional modeling method. In practice, instances of our framework lead to significant improvements in worst-group performance---a standard measure of group robustness--- across four diverse image understanding datasets. Notably, we achieve new state-of-the-art results on the WILDS leaderboards for both the Functional Map of the World (FMoW) \cite{fmow2018} and PovertyMap \cite{povertymap} datasets, improving worst-group accuracy on FMoW by a margin of +4\% and worst-group correlation coefficient ($r$) on PovertyMap by a margin of +0.04. 
%
%
These improvements come without a significant drop in overall average performance, and, in several cases, our methods even improve average performance. Overall, our findings indicate that modeling latent domain assignment using location encoders is an effective way to learn from data with geographically varying subpopulations. 

%% file: sec/2_background.tex
\section{Background}
\label{sec:background}
In this section, we describe the problem of subpopulation shift and discuss relevant domain adaptation and location-aware modeling methods from prior work. In \cref{sec:method}, we will detail how our framework incorporates conditional modeling strategies and auxiliary domain prediction loss from these works and adapts them for the geospatial setting.
\subsection{Subpopulation Shift}
What goes wrong when learning a predictor $f: X \rightarrow Y$ using ERM in the midst of a subpopulation shift? As different domains may have different sets of features that are predictive of the target $y$, the predictor must learn to rely on the correct set of features for each domain without becoming dependent on spurious correlations. Standard ERM can fail in this setting by either underfitting to predictive features from under-represented domains in the training data or by overfitting to spurious features from over-represented domains. In both cases, the result is a predictor that performs significantly better on some domains than others at inference time. This bias is problematic when the mixture of domains at inference time is significantly different from what is seen during training or when the bias can have unfair or harmful consequences when deployed \citep{groupdro}.

Many approaches to learning a predictor that is robust to subpopulation shift fall into one of three categories. The first seeks to strengthen the reliance of the predictor on domain-specific predictive features by using domain labels to condition the hypothesis space. Examples include feature-wise modulation \citep{domain-film} and mixture of experts models \citep{d3g}. The second seeks to reduce the reliance of the predictor on spurious correlations by learning features that satisfy some invariance condition across domains \citep{deepcoral, irm, dann}. Note that some recent approaches are more targeted and can be considered part of both categories \citep{lisa, targetedaug}. The third category consists of distributionally robust optimization (DRO) approaches that aim to minimize risk according to an estimated worst-case data distribution \citep{groupdro}. In this work, we focus on a subset of approaches from the first category.


\subsection{Conditional Modeling for Domain Adaptation}
\label{background:conditional_modeling}
In this section, we describe the conditional modeling methods for domain adaptation that we build on in this work. We focus on conditional modeling because spatial context is often readily available as metadata and is a natural signal to condition on.
\subsubsection{Domain Conditional Predictors (DCP)} The key idea behind domain conditional predictors \citep{domain-film} is to learn a latent representation of each domain and to condition the prediction of the target $y$ on these domain representations using feature-wise linear modulation (FiLM) \citep{film}. The latent domain representations are obtained from an intermediate layer in a neural network that is trained to predict the domain label directly from the input $x$. This network is trained jointly with the predictor $f$ and discarded at inference time. The DCP loss function consists of two terms: the task prediction (TP) loss, and the domain prediction (DP) loss weighted by a hyperparameter $\alpha$:
$\mathcal{L}_{TP} + \alpha \mathcal{L}_{DP}$.
Our proposed framework builds on the ideas of using a domain prediction loss and conditioning on a main task predictor, but adapts them for the geospatial setting.

\subsubsection{Domain Distances for OOD Generalization (D\textsuperscript{3}G)} Given domain labels at training time, D\textsuperscript{3}G \citep{d3g} trains separate prediction heads for each domain in the training data, and jointly trains another model to predict pairwise similarities between domains, called domain relations, from metadata. At inference time, the predictions output from various heads $j$ are weighted by their relation to the input domain $i$, calculated as an interpolation of pre-defined fixed and learnable domain relations. The learnable domain relation predictor is:
\begin{equation}
    \label{eq:d3g}
    \frac{1}{R} \sum_{r=1}^R S_C \left( ~ w_r \cdot g(m_i), ~~ w_r \cdot g(m_{\text{head \textit{j}}})~\right)
\end{equation}
where $S_C$ denotes cosine similarity, $m_i$, $m_{\text{head \textit{j}}}$ is metadata representing the domain of input $i$ and head $j$ respectively, ${w_r}$ is a set of learnable vectors, and $g$ is a neural network. 

\subsection{Geo-Aware Modeling and Location Encoders}
\label{sec:background_geoaware_modeling}
The goal of geo-aware modeling is to explicitly incorporate geographic information into the model design to improve its ability to learn spatial correlations. Geo-aware modeling is typically achieved using learning-friendly, high-dimensional representations of geographic location through location embedding models of the form $\ell(\phi, \lambda) \rightarrow \mathbb{R}^d$, where $\phi, \lambda$ denote geographic coordinates. The choice of location featurization method can be either discrete or continuous along the surface of the Earth. Examples of discrete featurization methods include grid-based methods \cite{feifei2015, space2vec} and satellite imagery-based methods \citep{rolf2021mosaiks, alphaearth}. In our work, we focus on continuous featurization methods, which we refer to as location encoders. 

There are several approaches to continuous featurization of location, many of which are benchmarked by TorchSpatial \citep{torchspatial}. Recently, several approaches to large-scale pre-training of such location encoders have also been proposed \citep{dollinger2025climplicit, klemmer2023satclip, mai2023csp, geoclip}. Whether pre-trained or trained from scratch, location encoders have proven to be useful for improving overall model performance on a variety of prediction tasks, including geo-localization \citep{geoclip}, species distribution modeling \citep{wrap}, image classification \citep{torchspatial}, regression \citep{klemmer2023satclip}, and super-resolution \citep{gan2025}. 
In our work, we build on two common methods of using location encoders in image understanding models: feature-level concatenation, a common multi-modal fusion technique \cite{baltrusaitis2019multimodal} that has been successfully applied to image and location encoder features \cite{klemmer2023satclip, geoclip}, and Geo Priors, which we describe next.

\subsubsection{Geo Priors}
First proposed in \cite{wrap}, the central idea behind Geo Priors is to encode spatial context in an image classification model as a Bayesian prior. The probability $P(y|l)$ of observing a class $y \in \{1, ..., C\}$ at a location $l = (\phi, \lambda)$ is modeled as $\sigma(h^{(y)}(\ell(\phi, \lambda)))$ where $\sigma$ is the sigmoid function, $\ell$ is a location encoder, and $h$ is a neural network with $h(\ell(\phi, \lambda)) \in \mathbb{R}^C$ and $h^{(y)}(\ell(\phi, \lambda))$ used to denote the $y$-th component of $h(\ell(\phi, \lambda))$. The network $h(\ell(\phi, \lambda))$ is trained using negative log loss:
\begin{align*}\mathcal{L}_{\text{GeoPrior}}(h(\ell(\phi, \lambda), y)) = \ & C \log\left(\sigma(h^{(y)}(\ell(\phi, \lambda)))\right) \\ + &\quad \sum_{i \neq y} \log \left(1 - \sigma(h^{(i)}(\ell(\phi, \lambda)))\right)~.
\end{align*}
For an image $x$, the probability $P(y|x)$ is modeled using a neural network $f(x)$, either trained separately or jointly with $h(\ell(\phi, \lambda))$. The final predicted class $y$ is then:
$$\text{argmax}_i ~ f^{(i)}(x) ~\sigma(h^{(i)}(\ell(\phi, \lambda)))~.$$



%% file: sec/3_method.tex
\begin{figure*}
\centering
    \includegraphics[width=1\textwidth]{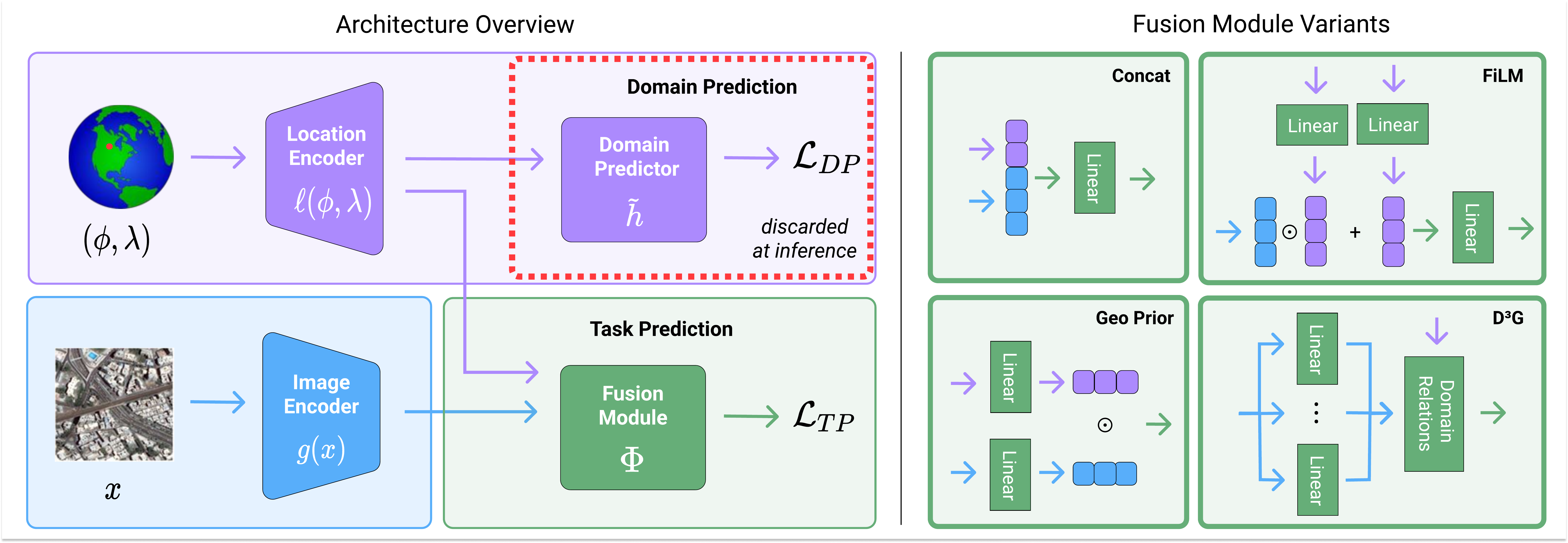}
  \caption{Illustration of our proposed framework. The three primary components are shown on the left: the image encoder, location encoder trained with domain prediction, and fusion module. The domain predictor, in dashed red outline, is discarded at inference time. In this work, we experiment with four options for the fusion module, shown on the right: feature concatenation, Geo Priors, FiLM, and D\textsuperscript{3}G.}
  \label{fig:main}
\end{figure*}

\section{Method}
\label{sec:method}

\newcommand{\dpweight}{\alpha}

In this section, we describe our proposed framework for improved robustness to geographically-induced subpopulation shift, which is depicted in Figure \ref{fig:main}.

\subsection{Overview}


We assume access to a dataset $\{(x_i, y_i, \phi_i, \lambda_i, d_i)\}$ where $x_i$ is an image, $y_i$ is the target, $(\phi_i, \lambda_i)$ are geographic coordinates associated with $x_i$, and $d_i$ is a geographic domain label (e.g. continents or biomes). We want to train a predictor $f$ on this dataset given a loss function $\mathcal{L}_{TP}$, called the task prediction loss (e.g. mean squared error or cross-entropy loss). We are interested in training $f$ to achieve low loss on a held-out dataset, not only overall across the entire set but also individually on each domain $d$. We propose a predictor of the form: $$f(x, \phi, \lambda) = \Phi(g(x), \ell(\phi, \lambda))$$
where $g$ is an image encoder, $\ell$ is a location encoder, and $\Phi$ is a predictor that fuses the image and location encoder outputs. We then propose to jointly train $f$ with a separate network $\Tilde{h}$ using a two-term loss function:
$$\mathcal{L}_{TP}(f(x, \phi, \lambda), y) + \dpweight \mathcal{L}_{DP}(\Tilde{h}(\ell(\phi, \lambda)), d)~.$$

Here, $d$ is the discrete domain label, $\Tilde{h}$ is a predictor for the domain, and $\mathcal{L}_{DP}$ is an auxiliary loss applied to the domain predictor. The domain prediction loss is weighted by a hyperparameter $\dpweight$. Note that domain labels $d$ are only required for training and $\Tilde{h}$ is discarded at inference time; its purpose is only to guide the training of the location encoder.  

The flexible design of our framework supports many different options for each of its components: the location encoder $\ell$ with domain predictor $\Tilde{h}$ and importance $\dpweight$, and the fusion module $\Phi$. In the next sections, we discuss some available choices for each component. Note that with $\dpweight = 0$, we recover a predictor that fuses location and image inputs but is trained to minimize only $\mathcal{L}_{TP}$, which may be useful if domain labels are not available at training time. Finally, while we focus on image understanding, we note that our framework is agnostic to the choice of $g$ and $\mathcal{L}_{TP}$ and can be applied to other modalities such as text or audio. 


\subsection{Location Encoder and Domain Prediction}

The location encoder $\ell$ used for conditional modeling and input to the domain predictor can be trained from scratch or initialized from a pre-trained encoder. If using a pre-trained encoder, it can be trained by either keeping the original weights frozen and passing the output to a trainable network, or by fine-tuning end-to-end. In this work, we primarily experiment with two choices of location encoder:


\begin{itemize}
    \item \textbf{WRAP} consists of a simple non-parametric sine-cosine encoder on top of which a fully connected network $F$ is trained from scratch. Following \cite{wrap} and \cite{sphere2vec}, the WRAP location encoder is defined as
    $\ell_{\text{WRAP}}(\phi, \lambda) = F( \left[ ~\sin(\phi) ~ \cos(\phi) ~ \sin(\lambda) ~ \cos(\lambda)~\right]~).$
    \item \textbf{GeoCLIP} \cite{geoclip} is a location encoder based on random Fourier features (RFF) pre-trained on image-location pairs from Flickr. We construct the location encoder $l_{\text{GeoCLIP}}$ as a fully connected network $F$ trained on top of frozen GeoCLIP features, as prior work has found that unfreezing location encoder weights in a similar setting can lead to poor generalization \cite{rao2025using}.
\end{itemize}  

Any suitable classifier can be used for the domain predictor $\Tilde{h}$. In our experiments, we use a single linear layer. The domain prediction loss is chosen to be standard cross entropy loss, weighted with $\dpweight = 0.2$. 

\subsection{Fusion Module}
We experiment with four different approaches to fusing image and location features which are inspired by prior works outlined in \Cref{background:conditional_modeling,sec:background_geoaware_modeling}. 
\begin{itemize}
\item \textbf{Concatenate.} Feature-level concatenation is a well-known multi-modal fusion technique \cite{baltrusaitis2019multimodal}. For this fusion approach, we use a linear layer $h$ as the main task predictor applied to concatenated image and location features:
$$\Phi(g(x), \ell(\phi, \lambda)) = h(\left[ ~ g(x) ~~\ell(\phi, \lambda) ~ \right])~.$$
\item \textbf{FiLM.} Inspired by domain conditional predictors (DCP) \citep{domain-film}, we use a single FiLM layer \cite{film} to condition image features on location embeddings. The fusion module is:
$$\Phi(g(x), \ell(\phi, \lambda)) = h\Big(\gamma(\ell(\phi, \lambda)) \odot g(x) + \beta(\ell(\phi, \lambda))\Big)$$
where $h$, $\gamma$, and $\beta$ are linear layers and $\odot$ is a Hadamard product.

\item \textbf{Geo Priors.} In the Geo Priors approach \cite{wrap} to fusing image and location features, the fusion module is given by:
$$\Phi(g(x), \ell(\phi, \lambda)) = h_\text{image}(g(x)) \cdot h_\text{loc}(\ell(\phi, \lambda))$$

where $h_\text{image}$ and $h_\text{loc}$ are linear layers.

\item \textbf{D\textsuperscript{3}G.} 
We adapt the original D\textsuperscript{3}G \cite{d3g} method to use the output of a location encoder $\ell$ instead of the domain metadata $m_i$ as input to the domain relation predictor in \cref{eq:d3g}. Note that $m_{\text{head \textit{j}}}$ and its encoding are unchanged. The separate prediction heads $h_j$ are linear layers. Our proposed location-aware D\textsuperscript{3}G fusion module is:
$$\Phi(g(x), \ell(\phi, \lambda)) = \sum_{j=1}^D \beta_j(\ell(\phi, \lambda)) ~ h_j(g(x))$$
where $\beta_j(\ell(\phi, \lambda))$ is the learnable domain relation between location $(\phi, \lambda)$ and domain $j$.
\end{itemize}

%% file: sec/4_experiments.tex
\section{Experiments}
\label{sec:experiments}
\input{sec/table2}

\subsection{Datasets}
We conduct experiments on four image understanding datasets: two from the WILDS \cite{wilds} benchmark, and two from the TorchSpatial \citep{torchspatial} benchmark. We have chosen these datasets to cover a broad range of tasks and settings.

\begin{itemize}
\item \textbf{WILDS-FMoW.} Functional Map of the World (FMoW) \cite{fmow2018} is a land-use classification dataset from RGB satellite imagery, covering over 200 countries and 62 classes. We use domain labels provided by the WILDS benchmark, where each continent is treated as a different domain. 
\item \textbf{WILDS-PovertyMap.} PovertyMap is a dataset of multi-spectral Sentinel-2 imagery for asset wealth prediction, which is a regression task. We use domain labels provided by WILDS corresponding to whether the image captures an urban or rural area.
\item \textbf{iNat-Biomes.} iNaturalist 2018 \cite{inat} is a species recognition dataset consisting of outdoor imagery from around the globe, labeled with one of 8,142 classes. Since the TorchSpatial benchmark does not provide domain labels, we derive them from geographic coordinates and ecoregion data from \cite{biomesdata}. Each image is assigned a domain corresponding to one of 14 biome classes. We use the ``Biomes" suffix to distinguish our dataset from the original TorchSpatial iNat2018 dataset, since we discarded roughly 10\% of the data from both train and val sets due to not corresponding to any of the 14 biome classes. 
\item \textbf{TorchSpatial-YFCC.} YFCC100M-GEO100 is a geo-tagged subset of the Yahoo Flickr Creative Commons 100M dataset \cite{yfcc100mgeo100} where the task is classification over 100 classes. In this subset, all images are collected in the United States. We use data from Natural Earth (public domain) to derive binary urban/rural domain labels. 
\end{itemize}

We evaluate the WILDS datasets on their official OOD test sets which exhibit a significant subpopulation shift by design. For iNat-Biomes, as with iNat2018, the val and test sets are the same \cite{torchspatial}, and the differences in biome distributions across train and test sets induces a subpopulation shift. The YFCC dataset is a slightly different setting as the domains are more evenly distributed and the subpopulation shift is less pronounced. 

\input{sec/pareto_figure}

\subsection{Models and Baselines}
For each dataset, we experiment with the two choices of location encoders (WRAP, GeoCLIP) and four choices of fusion module (Concatenate, Geo Priors, FiLM, D\textsuperscript{3}G) discussed in \cref{sec:method}. For the image encoder, we fine-tune a CLIP ViT-L/14 \cite{clip} for FMoW, and we train a randomly-initialized multi-spectral ResNet-18 for PovertyMap as in \cite{wilds}. For iNat-Biomes and YFCC, we use the same setup from TorchSpatial \cite{torchspatial} where image predictions (for Geo Priors) and 2048-dimensional image features (for other fusion methods) are fixed and precomputed. For the location encoder, the network $F$ trained on top of WRAP and GeoCLIP features consists of 4 residual blocks, similar to \cite{torchspatial}.

We compare our framework against two categories of baselines: \textbf{1)} The location-free baselines consist of standard ERM and domain adaptation methods: invariance-based methods (IRM, CORAL), GroupDRO, and D\textsuperscript{3}G. With the exception of ERM, all baselines from this category rely on domain labels during training. \textbf{2)} The location-aware modeling baselines are well-studied approaches to fusing location and image features: feature-level concatenation and Geo Priors. Note that we cannot evaluate IRM and CORAL on TorchSpatial datasets since the image features are frozen, and because Geo Priors are only applicable to classification tasks, we cannot evaluate this method on PovertyMap. These entries are marked with ``---" in our tables.

\section{Results}


\subsection{Main Results}
Our main results can be found in Table \ref{tab:main_results}. For each dataset, we report overall average and worst group performance, measured using the Pearson correlation coefficient $r$ for PovertyMap and accuracy on other datasets. Each result is averaged across 3 random seeds, except for PovertyMap results which are averaged across 5 data folds following the official WILDS guidelines. 
Please refer to the Supplementary section for a detailed report of our experimental setup. 

We find that our proposed conditional modeling methods with domain prediction consistently achieve the best worst-group performance on all four datasets. In particular, the GeoCLIP variants of D\textsuperscript{3}G w/ DP and FiLM w/ DP achieve a new state-of-the-art result on the WILDS leaderboards for FMoW and PovertyMap respectively (the current top results are 51.8\% for FMoW and $r = 0.53$ for PovertyMap).\footnote{Accessed at \href{https://wilds.stanford.edu/leaderboard/}{https://wilds.stanford.edu/leaderboard/} on 11/14/2025.} Notably, these methods incur only a modest increase in training cost compared to standard ERM, increasing the number of trainable parameters from 428M to 429M for FMoW and from 11.2M to 12.3M for PovertyMap. A full comparison of model sizes is included in the Supplementary.



We also observe that location-aware modeling tends to outperform standard domain adaptation methods in both overall average and worst-group performance, and that applying domain prediction improves average performance in many cases. In fact, the average accuracies of both WRAP and GeoCLIP variants of Concat w/ DP on YFCC exceed the top TorchSpatial benchmark result reported in \citet{torchspatial}. While GeoCLIP consistently leads to better performance than WRAP, we observe that the best performing fusion method is not consistent across datasets---a question we leave for future work.


\subsection{Worst Group vs. Average Performance}
Past work has found that improvements to worst-group performance in the subpopulation shift setting may come at the expense of performance on other groups \cite{groupdro}. In this section, we examine the trade-off between worst-group and overall average performance by plotting the results for each dataset and location encoder from Table \ref{tab:main_results} in a scatter plot. The plots for most dataset/location encoder combinations are shown in Figure \ref{fig:pareto}, and the remaining plots, which show similar trends, can be found in the Supplementary. We find that on each dataset and for both choices of location encoder, our proposed methods for conditional modeling with domain prediction loss (shown with star markers in Figure \ref{fig:pareto})  not only achieve the top worst-group performance, but are also consistently on the Pareto frontier, indicating that they optimally trade off worst-group and average performance compared to the baselines. On the other hand, standard domain adaptation methods using domain labels with no geographic coordinates (shown with grey circle markers in Figure \ref{fig:pareto}) achieve the poorest trade-off in most cases. 

\subsection{Ablations}
\label{sec:ablations}
In this section, we report ablation results on the two central components of our proposed framework: the location encoder and domain prediction loss. To conduct ablations on the location encoder, for each fusion method we replace the location encoder with a domain encoder that simply maps each domain to a learnable embedding which is used for conditioning. We apply the domain prediction loss to the domain embeddings with $\alpha = 0.2$ to avoid collapse. The worst-group results are shown in Table \ref{tab:loc_enc_ablation} and the full set of results can be found in the Supplement. We find that the location-aware method outperforms its domain encoder variant in the majority of cases, indicating that modeling latent domain assignment as a continuous function of input is more effective than learning domain representations or relations from discrete labels. For ablation of the domain prediction loss, we report in Table \ref{tab:dp_ablation} the difference in worst-group performance when using the domain prediction loss compared to no domain prediction ($\alpha = 0$). The full set of results can be found in the Supplement. In the vast majority of cases, we observe that domain prediction improves worst-group performance across fusion methods, location encoders, and datasets.




%% file: sec/table2.tex
\begin{table*}[t]
\centering
\renewcommand{\arraystretch}{1.25}
\setlength{\tabcolsep}{6pt}
\footnotesize
\begin{tabular}{@{}p{1.5cm} l l cc cc cc cc@{}}
\toprule
\multirow{2}{*}{} &
\multirow{2}{*}{\textbf{Loc. Enc.}} &
\multirow{2}{*}{\textbf{Method}} &
\multicolumn{2}{c}{\textbf{FMoW}} &
\multicolumn{2}{c}{\textbf{PovertyMap}} &
\multicolumn{2}{c}{\textbf{iNat-Biomes}} &
\multicolumn{2}{c}{\textbf{YFCC}} \\
\cmidrule(lr){4-5} \cmidrule(lr){6-7} \cmidrule(lr){8-9} \cmidrule(lr){10-11}
 & & & \textbf{Avg} & \textbf{Worst} & \textbf{Avg} & \textbf{Worst} & \textbf{Avg} & \textbf{Worst} & \textbf{Avg} & \textbf{Worst} \\
\midrule

\multirow{5}{=}{\textbf{Baselines\\}}
& None & ERM        & \textbf{67.1 (0.1)} & 47.6 (0.9) & 0.78 (.02) & 0.45 (.03) & 58.8 (0.0) & 53.1 (0.5) & 50.7 (0.0) & 48.1 (0.1) \\
& None & IRM        & 64.7 (0.1) & 48.4 (1.3) & 0.77 (.02) & 0.41 (.04) & --- & --- & --- & --- \\
& None & CORAL      & 65.3 (0.3) & 49.4 (0.7) & 0.76 (.03) & 0.39 (.05) & --- & --- & --- & --- \\
& None & GroupDRO   & 65.0 (0.6) & 50.0 (1.2) & 0.77 (.02) & 0.41 (.04) & 58.5 (0.0) & 51.5 (0.9) & 51.2 (0.1) & 48.2 (0.1) \\
& None & D\textsuperscript{3}G        & 66.9 (0.2) & 50.4 (0.1) & 0.76 (.02) & 0.37 (.04) & 63.4 (0.3) & 52.3 (1.5) & 51.4 (0.0) & 48.2 (0.0) \\
\midrule
\multirow{4}{=}{\textbf{Baselines\\}}
& WRAP        & Concat & 66.9 (0.2) & 48.0 (2.0) & 0.78 (.02) & 0.48 (.04) & 71.7 (0.1) & 68.4 (0.3) & 51.5 (0.0) & 48.3 (0.1) \\
& WRAP        & Geo Priors    & 65.7 (0.2) & 48.1 (1.7) & ---\ & ---  & 72.5 (0.0) & 65.2 (0.2) & 50.5 (0.0) & 47.4 (0.0) \\
& GeoCLIP     & Concat & 66.0 (0.3) & 52.8 (1.0) & \underline{0.80 (.02)} & 0.53 (.03) & \textbf{74.6 (0.1)} & \underline{70.7 (0.6)} & \textbf{52.4 (0.0)} & \textbf{49.2 (0.1)} \\
& GeoCLIP     & Geo Priors    & 66.8 (0.2) & 49.6 (0.8) & ---\ & ---  & 72.8 (0.0) & 64.7 (0.6) & 50.9 (0.0) & 47.9 (0.0) \\
\midrule

\multirow{8}{=}{\textbf{Ours\\ (all w/ DP)}}
& WRAP        & Concat  & 66.8 (0.4) & 49.1 (1.6) & 0.80 (.02) & 0.52 (.04) & 73.1 (0.0) & 70.2 (0.2) & \underline{51.6 (0.1)} & \underline{48.4 (0.2)} \\
& WRAP        & Geo Priors & 65.7 (0.2) & 49.1 (2.0) & ---\ & ---  & 72.6 (0.03) & 65.5 (0.2) & 50.5 (0.0) & 47.4 (0.1) \\
& WRAP        & FiLM  & 66.9 (0.5) & 51.8 (2.1) & 0.80 (.02) & 0.52 (.04) & 61.5 (1.2) & 56.0 (1.1) & 49.7 (0.0) & 46.3 (0.1) \\
& WRAP        & D\textsuperscript{3}G          & \underline{66.9 (0.2)} & \underline{53.1 (0.7)} & 0.79 (.02) & 0.47 (.04) & 65.2 (1.6) & 47.0 (1.4) & 51.3 (0.1) & 48.2 (0.2) \\
& GeoCLIP     & Concat & 66.2 (0.2) & 52.0 (0.3) & 0.80 (.02) & \underline{0.54 (.02)} & \underline{73.6 (0.0)} & \textbf{71.1 (0.1)} & \textbf{52.4 (0.0)} & \textbf{49.2 (0.0)} \\
& GeoCLIP     & Geo Priors     & 66.8 (0.1) & 50.9 (1.1) & ---\ & --- & 72.6 (0.03) & 66.9 (0.5) & 50.9 (0.0) & 47.9 (0.0) \\
& GeoCLIP     & FiLM         & 66.9 (0.8) & 51.7 (2.0) & \textbf{0.82 (.02)} & \textbf{0.57 (.03)} & 53.1 (1.5) & 49.8 (1.6) & 49.7 (0.2) & 46.2 (0.2) \\
& GeoCLIP     & D\textsuperscript{3}G         & 66.6 (0.1) & \textbf{55.8 (0.7)} & 0.79 (.02) & 0.46 (.04) & 60.7 (2.7) & 55.1 (0.7) & 51.3 (0.0) & 48.2 (0.1) \\
\bottomrule
\end{tabular}
\caption{Our main experiment results. For each dataset, we report overall average and worst-group performance. All results are averaged over 5 data folds for PovertyMap and over 3 random seeds for all other datasets, with standard error reported in parentheses. Across all datasets, the top worst-group result is achieved by an instance of our proposed framework.}
\label{tab:main_results}
\end{table*}

%% file: sec/pareto_figure.tex
    

\begin{figure*}
    \centering
    \begin{subfigure}{0.3\textwidth}
        \centering
        \includegraphics[width=\linewidth]{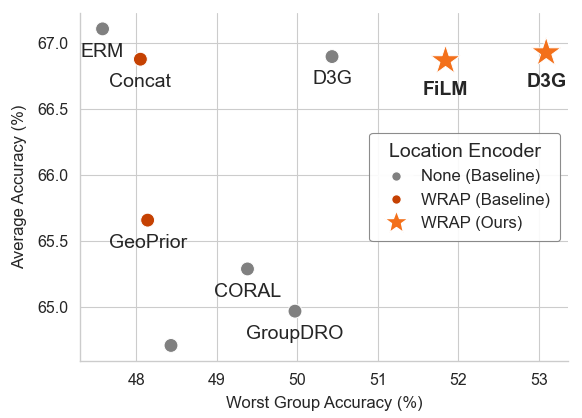}
        \caption{FMoW WRAP}
    \end{subfigure}
    \hspace{5pt}
    \begin{subfigure}{0.3\textwidth}
        \centering
        \includegraphics[width=\linewidth]{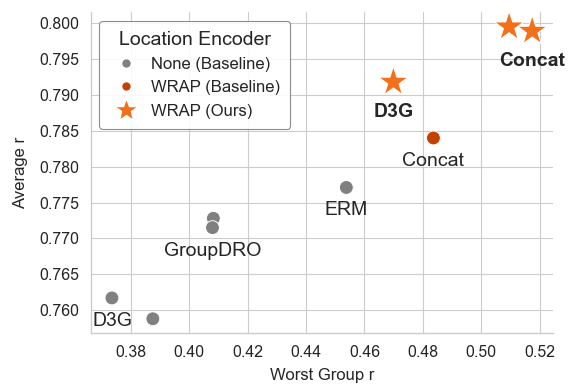}
        \caption{PovertyMap WRAP}
    \end{subfigure}
    \hspace{5pt}
        \begin{subfigure}{0.3\textwidth}
        \centering
        \includegraphics[width=\linewidth]{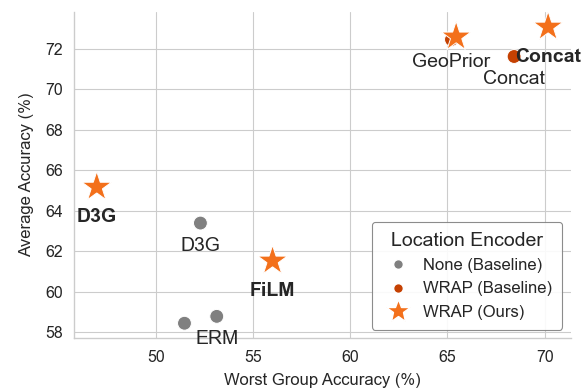}
        \caption{iNat-Biomes WRAP}
    \end{subfigure}
    \vspace{10pt}
    \hspace{5pt}
    \begin{subfigure}{0.3\textwidth}
        \centering
        \includegraphics[width=\linewidth]{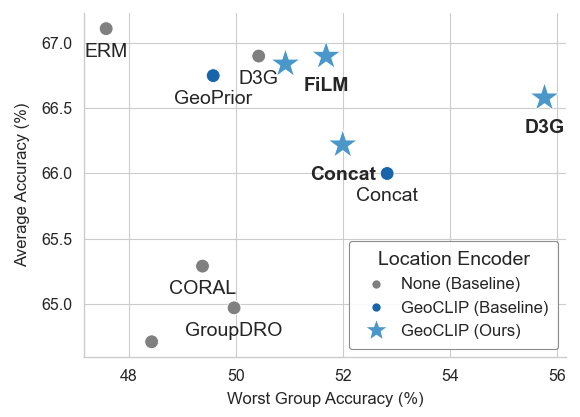}
        \caption{FMoW GeoCLIP}
    \end{subfigure}
    \begin{subfigure}{0.3\textwidth}
        \centering
        \includegraphics[width=\linewidth]{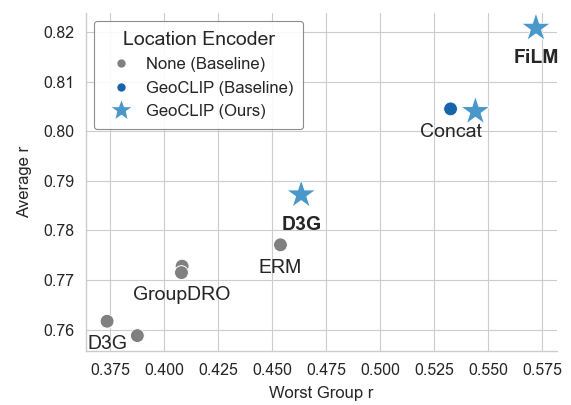}
    \caption{PovertyMap GeoCLIP}
    \end{subfigure}
    \hspace{5pt}
    \begin{subfigure}{0.3\textwidth}
        \centering
        \includegraphics[width=\linewidth]{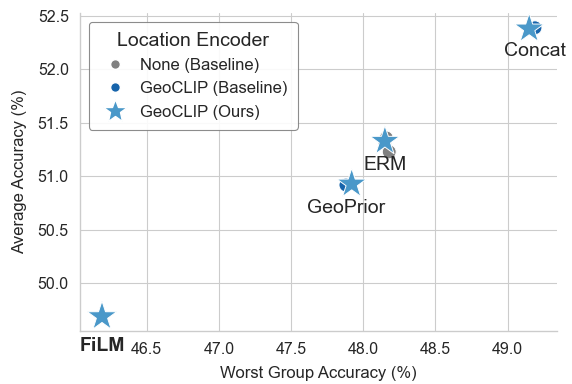}
        \caption{YFCC GeoCLIP}
    \end{subfigure}
    \caption{Scatter plots for different combinations of dataset and location encoder, each showing the overall average against worst group performance across different methods. Each result is averaged over the official 5 data folds for PovertyMap and over 3 random seeds for other datasets. Our proposed methods are consistently on the Pareto frontier in all plots.}
    \label{fig:pareto}
\end{figure*}

%% file: sec/5_discussion.tex
\input{sec/loc_enc_ablation_table}
\input{sec/dp_ablation_table}

\subsection{Additional Location Encoders}
To further validate our proposed framework, we conduct experiments on PovertyMap and iNat-Biomes with two additional location encoders: random Fourier features (RFF) trained from scratch, and pre-trained SatCLIP \cite{klemmer2023satclip}. The results can be found in Table \ref{tab:additional_loc_enc}. Regardless of the choice of location encoder, we consistently observe that instances of our proposed framework achieve the best worst-group performance. Interestingly, the best choice of fusion method for both datasets is consistent across all location encoders (including WRAP and GeoCLIP), though inconsistent across datasets as noted earlier.

\input{sec/additional_loc_enc_table}

\begin{figure*}
\centering
\includegraphics[width=1\textwidth]{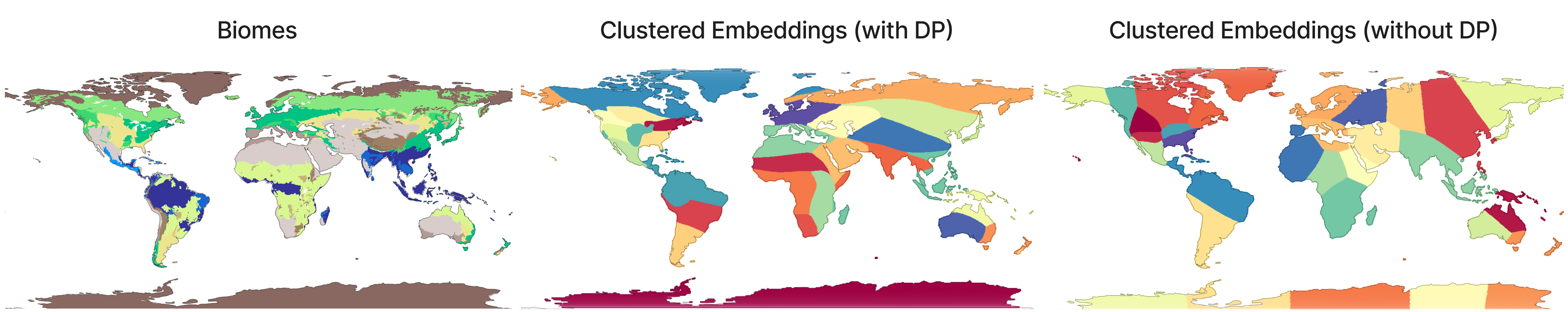}
  \caption{\textbf{Left}: Map of the 14 biomes used in the iNat-Biomes dataset. \textbf{Center}: Visualization of location embedding clusters obtained from a WRAP Concat model trained on iNat-Biomes with domain prediction weight $\alpha = 0.2$. Each color represents a different cluster. \textbf{Right}: Location embedding clusters using the same model trained without domain prediction. The domain prediction loss in our proposed framework allows the location embedding space to align more closely with the true spatial distribution of domains.}
  \label{fig:maps}
    \vspace{-0.05in}
\end{figure*}

\subsection{Visualizing Location Embedding Space}
\label{sec:viz}
In this section, we seek to better understand why domain prediction loss improves worst-group performance by visualizing the learned location embedding space. Specifically, we train two models on iNat-Biomes with a WRAP location encoder using the concatenate fusion method: one with domain prediction ($\alpha$ = 0.2) and another without ($\alpha =0$). We then sample 100k locations uniformly across the Earth's landmass and apply $k$-means clustering to the trained WRAP embeddings with $k=28$. We visualize the results on a map in Figure \ref{fig:maps}, where each cluster is associated with a separate color. For comparison, we show these results alongside the ground truth biome polygons.

We observe that the structure of the cluster map for the WRAP embeddings trained with domain prediction aligns closely in most regions with the true biome map. For example, the cluster map correctly depicts the boundaries between the grasslands and deserts in Africa (shown in lime green and light brown, respectively, in the true map) as well as the boundary between taiga (shown in bright green in the true map) and other biomes in the northern hemisphere. On the other hand, the cluster map without domain prediction exhibits a noticeably different structure that over-fragments each of these biomes. Overall, these visualizations confirm that applying an auxiliary domain prediction loss has the expected effect of guiding the location embedding space towards a structure that closely reflects the true spatial distribution of domains. This suggests, together with the results from our domain prediction loss ablation in Section \ref{sec:ablations}, that conditioning on location embeddings that capture the spatial distribution of domains is what leads to a better outcome for the worst-performing domain. Conversely, conditioning on a discrete domain signal alone is not enough to produce these improved outcomes, as evidenced by our location encoder ablations in \Cref{tab:loc_enc_ablation}. 

\subsection{Sensitivity to Domain Prediction Weight}
\begin{figure}
\centering
    \includegraphics[width=0.4\textwidth]{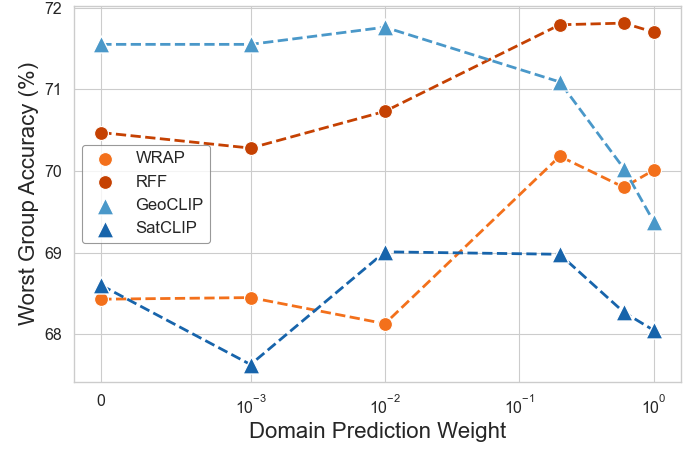}
  \caption{Worst group accuracy of the WRAP Concat models trained on iNat-Biomes for different values of the domain prediction weight $\alpha$. Each result is averaged over 3 random seeds.}
  \label{fig:sensitivity}
  \vspace{-.1in}
\end{figure}

In this section, we consider the effect of the domain prediction weight $\alpha$ on worst-group performance. In Figure \ref{fig:sensitivity}, we plot worst-group accuracy using the concatenate fusion method on iNat-Biomes across all four location encoders. We observe that for location encoders trained from scratch (WRAP, RFF), the worst-group accuracy follows an increasing trend as domain prediction weight grows, whereas for pre-trained location encoders (GeoCLIP, SatCLIP), a larger domain weight has the opposite effect. In fact, for these location encoders, setting the domain prediction weight too high degrades worst-group performance compared to not using domain prediction at all. Considering our findings from the visualizations in \Cref{sec:viz}, these results suggest that the effect of domain prediction on location encoders trained from scratch might differ from its effect on pre-trained encoders because the latter are likely to already possess some underlying useful structure. 



%% file: sec/loc_enc_ablation_table.tex
\begin{table}
\centering
\renewcommand{\arraystretch}{1.25}
\setlength{\tabcolsep}{4pt}
\footnotesize
\begin{tabular}{@{}p{1.1cm} l c c c c@{}}
\toprule
\textbf{Method} &
\textbf{Loc. Enc.} &
\textbf{FMoW} &
\textbf{PovertyMap} &
\textbf{iNat-Biomes} &
\textbf{YFCC} \\
\midrule

\multirow{3}{=}{Concat}
& None   & 50.9 & 0.40 & 49.9 & 48.3 \\
& WRAP   & $-1.8$ & $+0.12$ & $+20.3$ & $+0.1$ \\
& GeoCLIP& $+1.1$ & $+0.14$ & $+21.2$ & $+0.9$ \\
\midrule

\multirow{3}{=}{Geo Priors}
& None   & 50.0 & ---  & 43.7 & 47.1 \\
& WRAP   & $-0.9$ & ---  & $+21.8$ & $+0.3$ \\
& GeoCLIP& $+0.9$ & ---  & $+23.2$ & $+0.8$ \\
\midrule

\multirow{3}{=}{FiLM}
& None   & 50.0 & 0.43 & 52.1 & 46.9 \\
& WRAP   & $+1.8$ & $+0.08$ & $+3.9$ & $-0.6$ \\
& GeoCLIP& $+1.7$ & $+0.14$ & $-2.3$ & $-0.7$ \\
\midrule

\multirow{3}{=}{D\textsuperscript{3}G}
& None   & 51.7 & 0.46 & 44.9 & 48.3 \\
& WRAP   & $+1.4$ & $+0.01$ & $+2.1$ & $-0.1$ \\
& GeoCLIP& $+4.1$ & $+0.00$ & $+10.2$ & $-0.1$ \\
\bottomrule
\end{tabular}
\caption{Ablation results for the choice of location encoder within our  framework. Settings where location encoder = ``None" map discrete domain labels to latent representations before fusion; other numbers are additive differences from these baselines. }
\label{tab:loc_enc_ablation}
\vspace{-0.05in}
\end{table}






%% file: sec/dp_ablation_table.tex
\begin{table}[t]
\centering
\renewcommand{\arraystretch}{1.25}
\setlength{\tabcolsep}{4pt}
\footnotesize
\begin{tabular}{@{}l l c c c c@{}}
\toprule
\textbf{Method} & \textbf{Loc. Enc.} &
\textbf{FMoW} & \textbf{PovertyMap} &
\textbf{iNat-Biomes} & \textbf{YFCC} \\
\midrule

\multirow{2}{*}{Concat} 
& WRAP    & +1.1 & +0.04 & +1.8 & +0.1 \\
& GeoCLIP & --0.8 & +0.01 & +0.4 & 0.0 \\[3pt]
\midrule

\multirow{2}{*}{Geo Priors}
& WRAP    & +1.0 & ---   & +0.3 & 0.0 \\
& GeoCLIP & +1.3 & ---   & +2.2 & 0.0 \\[3pt]
\midrule

\multirow{2}{*}{FiLM}
& WRAP    & +2.4 & +0.02 & +3.7 & +0.1 \\
& GeoCLIP & +1.8 & 0.00  & +5.9 & --0.1 \\[3pt]
\midrule

\multirow{2}{*}{D$^3$G}
& WRAP    & --1.0 & +0.01 & +0.2 & --0.1 \\
& GeoCLIP & +1.8  & +0.01 & +1.5 & 0.0 \\
\bottomrule
\end{tabular}
\caption{Ablation results showing the additive difference in worst-group performance due to applying an auxiliary DP loss.}
\label{tab:dp_ablation}
\end{table}

%% file: sec/additional_loc_enc_table.tex
\begin{table}
\centering
\renewcommand{\arraystretch}{1.25}
\setlength{\tabcolsep}{5pt} 
\footnotesize
\begin{tabular}{@{}p{1.6cm} l l c c@{}}
\toprule
&
\textbf{Loc. Enc.} &
\textbf{Method} &
\textbf{PovertyMap} &
\textbf{iNat-Biomes}\\
\midrule

\multirow{4}{=}{\textbf{Baselines}}
& RFF & Concat & 0.53 (0.05) & \underline{70.5 (0.2)}\\
& RFF & Geo Priors & --- & 65.7 (0.8) \\
&SatCLIP & Concat & 0.48 (0.03) & 68.6 (0.5) \\
&SatCLIP & Geo Priors & --- & 64.0 (0.2) \\
\midrule

\multirow{8}{=}{\textbf{Ours\\(all w/ DP)}}
& RFF & Concat & 0.54 (0.03) & \textbf{71.8 (0.2)} \\
& RFF & Geo Priors & --- & 65.5 (0.6) \\
& RFF & FiLM & \textbf{0.56 (0.03)} & 55.8 (0.6) \\
& RFF & D\textsuperscript{3}G & 0.46 (0.04) & 54.7 (7.0) \\
&SatCLIP & Concat & 0.50 (0.03) & 69.0 (0.4)\\
&SatCLIP & Geo Priors & --- & 64.2 (0.4)\\
&SatCLIP & FiLM & \underline{0.55 (0.05)} & 64.3 (0.6) \\
&SatCLIP & D\textsuperscript{3}G & 0.48 (0.04) & 62.6 (0.5) \\
\bottomrule
\end{tabular}
\caption{Worst-group performance results with additional location encoders: RFF and SatCLIP. Regardless of the choice of location encoder, our methods achieve the top worst-group performance.}
\label{tab:additional_loc_enc}
\vspace{-.05in}
\end{table}

%% file: sec/6_conclusion.tex
\section{Conclusion}
\label{sec:conclusion}
Subpopulation shift can pose a major obstacle to the deployment of global-scale deep learning models, particularly in applications where training data is spatially imbalanced. In this paper, we propose a general modeling framework for improving the worst-group performance of such models. Our framework leverages location encoders trained with an auxiliary domain prediction loss to model continuous, latent domain assignment. Since location encoders are relatively lightweight, they provide an efficient way of learning these domain latents. We demonstrate that instances of our proposed framework achieve significant improvements in worst-group performance and that they optimally trade off worst-group and overall average performance compared to existing baselines. 
Future work may involve applying our framework to other prediction tasks or understanding the suitability of different fusion methods within our framework for different datasets, group splits, and tasks.




%% file: sec/X_suppl.tex
\clearpage
\setcounter{page}{1}
\maketitlesupplementary

\section{Dataset Details}
In this section, for each dataset used in this paper, we report the total size and domain distribution within each split. 
\paragraph{WILDS-FMoW} We use the default WILDS training set and OOD val and test sets with continent domain labels:
\begin{table}[h]
\centering
\renewcommand{\arraystretch}{1.2}
\setlength{\tabcolsep}{5pt}
\footnotesize
\begin{tabular}{lcccccc}
\toprule
\multirow{2}{*}{\textbf{Group}} & \multicolumn{2}{c}{\textbf{Train}} &
\multicolumn{2}{c}{\textbf{Val}}  &
\multicolumn{2}{c}{\textbf{Test}} \\
\cmidrule(lr){2-3}
\cmidrule(lr){4-5}
\cmidrule(lr){6-7}
&
\textbf{Count} & \textbf{\%} &
\textbf{Count}   & \textbf{\%} &
\textbf{Count}  & \textbf{\%} \\
\midrule
Asia     & 17,809 & 23\% & 4,121 & 21\% & 4,963 & 22\% \\
Europe   & 34,816 & 45\% & 7,732 & 39\% & 5,858 & 26\% \\
Africa   & 1,582  & 2\%  & 803  & 4\%  & 2,593 & 12\% \\
Americas & 20,973 & 27\% & 6,562 & 33\% & 8,024 & 36\% \\
Oceania  & 1,641  & 2\%  & 693  & 3\%  & 666  & 3\% \\
Other    & 42    & 0\%  & 4    & 0\%  & 4    & 0\% \\
\midrule
\textbf{Total:} &
\textbf{76,863} & \textbf{100\%} &
\textbf{19,915} & \textbf{100\%} &
\textbf{22,108} & \textbf{100\%} \\
\bottomrule
\end{tabular}
\caption{Continent distribution in WILDS-FMoW train, OOD validation, and OOD test sets.}
\end{table}

\paragraph{WILDS-PovertyMap} We use the default WILDS training set and OOD val and test sets with urban/rural domain labels:
\begin{table}[h]
\centering
\renewcommand{\arraystretch}{1.2}
\setlength{\tabcolsep}{6pt}
\footnotesize
\begin{tabular}{lcccccc}
\toprule
\multirow{2}{*}{\textbf{Group}} &
\multicolumn{2}{c}{\textbf{Train}} &
\multicolumn{2}{c}{\textbf{Val}}  &
\multicolumn{2}{c}{\textbf{Test}} \\
\cmidrule(lr){2-3}
\cmidrule(lr){4-5}
\cmidrule(lr){6-7}
& \textbf{Count} & \textbf{\%} & \textbf{Count} & \textbf{\%} & \textbf{Count} & \textbf{\%} \\
\midrule
Rural & 6,410 & 66\% & 2,647 & 67\% & 2,455 & 62\% \\
Urban & 3,368 & 34\% & 1,316 & 33\% & 1,473 & 37\% \\
\midrule
\textbf{Total:} &
\textbf{9,778} & \textbf{100\%} &
\textbf{3,963} & \textbf{100\%} &
\textbf{3,928} & \textbf{100\%} \\
\bottomrule
\end{tabular}
\caption{Urban/Rural distribution in WILDS-PovertyMap train, OOD validation, and OOD test sets.}
\end{table}


\paragraph{iNat-Biomes} iNat-Biomes is a subset of the TorchSpatial-iNat2018 \cite{torchspatial} dataset. About 10\% of datapoints were removed because they did not correspond to any of the 14 biome classes. The biome distribution across data splits in iNat-Biomes is shown in \cref{tab:supp_biome-dist}

\paragraph{TorchSpatial-YFCC} The train, validation, and test splits are unmodified from the TorchSpatial YFCC dataset \cite{torchspatial}:

\begin{table}[h]
\centering
\renewcommand{\arraystretch}{1.2}
\setlength{\tabcolsep}{6pt}
\footnotesize
\begin{tabular}{lcccccc}
\toprule
\multirow{2}{*}{\textbf{Group}}  &
\multicolumn{2}{c}{\textbf{Train}} &
\multicolumn{2}{c}{\textbf{Val}}  &
\multicolumn{2}{c}{\textbf{Test}} \\
\cmidrule(lr){2-3}
\cmidrule(lr){4-5}
\cmidrule(lr){6-7}
& \textbf{Train} & \textbf{\%} &
\textbf{Val}   & \textbf{\%} &
\textbf{Test}  & \textbf{\%} \\
\midrule
Rural & 33,360 & 50\% & 2,156 & 49\% & 9,044 & 51\% \\
Urban & 33,379 & 50\% & 2,293 & 51\% & 8,754 & 49\% \\
\midrule
\textbf{Total:} &
\textbf{66,739} & \textbf{100\%} &
\textbf{4,449}  & \textbf{100\%} &
\textbf{17,798} & \textbf{100\%} \\
\bottomrule
\end{tabular}
\caption{Urban/Rural distribution in TorchSpatial-YFCC train, validation, and test sets.}
\end{table}

\begin{table*}[h]
\centering
\renewcommand{\arraystretch}{1.25}
\setlength{\tabcolsep}{4pt}
\footnotesize
\begin{tabular}{p{7cm}cccccc}
\toprule
\multirow{2}{*}{\textbf{Group}} &
\multicolumn{2}{c}{\textbf{Train}} &
\multicolumn{2}{c}{\textbf{Val}}  &
\multicolumn{2}{c}{\textbf{Test}} \\
\cmidrule(lr){2-3}
\cmidrule(lr){4-5}
\cmidrule(lr){6-7}
& \textbf{Count} & \textbf{\%} &
\textbf{Count}   & \textbf{\%} &
\textbf{Count}  & \textbf{\%} \\
\midrule

1. Tropical \& Subtropical Moist Broadleaf Forests 
 & 13,521 & 3\% & 1,530 & 7\% & 1,530 & 7\% \\

2. Tropical \& Subtropical Dry Broadleaf Forests
 & 10,204 & 3\% & 872 & 4\% & 872 & 4\% \\

3. Tropical \& Subtropical Coniferous Forests
 & 7,222 & 2\% & 557 & 3\% & 557 & 3\% \\

4. Temperate Broadleaf \& Mixed Forests
 & 125,749 & 31\% & 6,817 & 31\% & 6,817 & 31\% \\

5. Temperate Conifer Forests
 & 41,480 & 10\% & 2,243 & 10\% & 2,243 & 10\% \\

6. Boreal Forests/Taiga
 & 1,237 & 0\% & 113 & 1\% & 113 & 1\% \\

7. Tropical \& Subtropical Grasslands, Savannas \& Shrublands
 & 16,742 & 4\% & 1,140 & 5\% & 1,140 & 5\% \\

8. Temperate Grasslands, Savannas \& Shrublands
 & 81,881 & 20\% & 137 & 1\% & 137 & 1\% \\

9. Flooded Grasslands \& Savannas
 & 2,163 & 1\% & 69 & 0\% & 69 & 0\% \\

10. Montane Grasslands \& Shrublands
 & 1,279 & 0\% & 3,430 & 16\% & 3,430 & 16\% \\

11. Tundra
 & 762 & 0\% & 151 & 1\% & 151 & 1\% \\

12. Mediterranean Forests, Woodlands \& Scrub
 & 70,319 & 17\% & 2,786 & 13\% & 2,786 & 13\% \\

13. Deserts \& Xeric Shrublands
 & 29,631 & 7\% & 1,790 & 8\% & 1,790 & 8\% \\

14. Mangroves
 & 2,437 & 1\% & 149 & 1\% & 149 & 1\% \\

\midrule
\textbf{Total:} &
\textbf{404,627} & \textbf{100\%} &
\textbf{21,784}  & \textbf{100\%} &
\textbf{21,784}  & \textbf{100\%} \\
\bottomrule
\end{tabular}
\caption{Biome distributions in iNat-Biomes. As with TorchSpatial-iNat2018, the test and validation sets are the same.}
\label{tab:supp_biome-dist}
\end{table*}

\section{Model Details and Sizes}
\input{sec/sup/model_sizes}

\section{Training Hyperparameters}
We report training hyperparameters and design choices specific to each dataset (e.g. batch size, number of epochs) and specific to each model component and fusion method.

\paragraph{WILDS-FMoW} Each model is trained for 5 epochs and we select the model checkpoint with lowest validation loss across epochs. We train with batch size 16 and use the Adam optimizer with initial learning rate $10^{-4}$ that decays by a factor of 0.96 each epoch. The CLIP ViT-L/14 backbone is finetuned with the same optimizer and learning rate schedule but with initial learning rate $10^{-5}$. We use gradient accumulation to achieve an effective batch size of 64. Finally, we normalize all images using ImageNet mean and standard deviation, and apply random horizontal flip. 

\paragraph{WILDS-PovertyMap} We follow a training setup identical to \cite{wilds}, where each model is trained for 200 epochs and we select the model checkpoint with highest validation $r$ across epochs. We train with batch size 64 using Adam with initial learning rate $10^{-3}$ that decays by a factor of 0.96 each epoch. As in \cite{wilds}, we apply random horizontal and vertical flip and color jitter to each image.

\paragraph{iNat-Biomes} Each model is trained for 50 epochs and we select the model checkpoint with highest validation accuracy across epochs. We train with batch size of 1024 using Adam with initial learning rate $10^{-3}$ that decays by a factor of 0.96 each epoch. We use the image predictions and 2048-dimensional frozen image features from \cite{torchspatial} exactly as-is, without modification.

\paragraph{TorchSpatial-YFCC} Each model is trained for 100 epochs and we select the model checkpoint with highest validation accuracy across epochs. We train with batch size of 2048 using Adam with initial learning rate $10^{-4}$ that decays by a factor of 0.96 each epoch. As with iNat-Biomes, we use the image predictions and 2048-dimensional frozen image features from \cite{torchspatial} exactly as-is, without modification.

\paragraph{D\textsuperscript{3}G} There are two training hyperparameters used in D\textsuperscript{3}G, denoted $\lambda$ and $\beta$ in the original paper \cite{d3g}. D\textsuperscript{3}G has two loss terms: the task prediction loss (denoted $\mathcal{L}_{pred}$ in the original paper) computed from the prediction head for the current domain, and the consistency loss ($\mathcal{L}_{rel}$) that is computed from all other prediction heads whose predictions are weighted by their relation to the current domain. The final loss is of the form: $\mathcal{L}_{pred} + \lambda \mathcal{L}_{rel}$.
The $\beta$ hyperparameter is used to average fixed and learned domain relations (see \cite{d3g} for more details). We use a value of $\lambda = 0.5$ and $\beta = 0.8$ for all experiments.

\paragraph{Domain Prediction} The linear layer we use for domain prediction in all experiments is trained with Adam and step decay learning rate schedule (decayed by a factor of 0.96 per epoch) but with initial learning rate 0.1 times the initial learning rate of the prediction head. We tuned the domain prediction weight $\alpha$ on a subset of models, experimenting with values in $\{0.001, 0.01, 0.1, 0.2\}$, and used $\alpha = 0.2$ for most other experiments as it generally performed best. 

\section{Additional Results}
\begin{figure*}
\centering
\begin{subfigure}{0.4\textwidth}
    \centering
    \includegraphics[width=\textwidth]{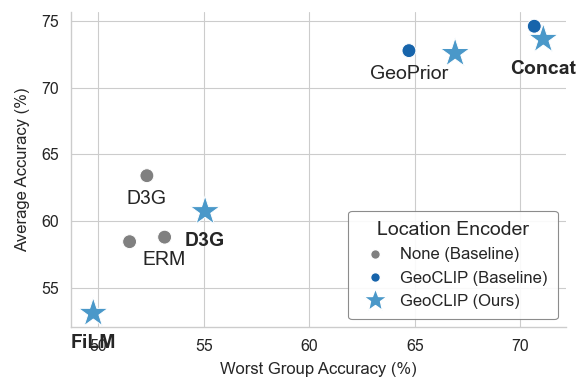}
    \caption{iNat-Biomes GeoCLIP}
    \label{supp_pareto_1}
\end{subfigure}
    \hspace{20pt}
\begin{subfigure}{0.4\textwidth}
    \centering
    \includegraphics[width=\textwidth]{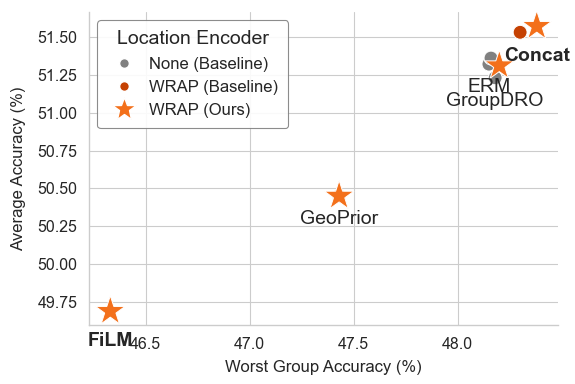}
    \caption{YFCC WRAP}
    \label{supp_pareto_2}
\end{subfigure}
\caption{Overall average accuracy vs. worst group accuracy. Each result is averaged over 3 random seeds.}
\label{fig:pareto_combined}
\end{figure*}

\paragraph{Additional Main Results Plots}
We include the two scatter plots showing overall average against worst group performance that were missing from \cref{fig:pareto} due to space constraints. They are shown in Figures \ref{supp_pareto_1} and \ref{supp_pareto_2}. These additional plots are consistent with those in the main section: we observe that our instances of our proposed framework outperform all other methods on worst-group performance, and that these instances lead to significant gains in overall average performance as well compared with other baselines.


\paragraph{Full Ablation Results} We report full ablation results (i.e. including overall average and worst group performance) for the auxiliary domain prediction loss in \cref{tab:dp_ablation_full} and for the location encoder in \cref{tab:loc_enc_ablation_full}. 
\input{sec/sup/dp_ablation_full}
\input{sec/sup/loc_enc_ablation_full}

\paragraph{Full RFF/SatCLIP Results} In \cref{tab:additional_loc_enc_full} we report full results (i.e. including overall average and worst group performance) on PovertyMap and iNat-Biomes using additional location encoders: RFF and SatCLIP.
\input{sec/sup/additional_loc_encoders_full}

%% file: sec/sup/model_sizes.tex
In this section we report model architecture details and sizes for each location encoder. In \cref{tab:supp_sizes_fmow,tab:supp_sizes_poverty,tab:supp_sizes_inat,tab:supp_sizes_yfcc} we report total model sizes for each combination of location encoder and fusion method. In \cref{tab:supp_dp_sizes} we report domain predictor sizes. 

\subsection{Location Encoders}
We train a ResNet consisting of 4 residual blocks similar to $\textbf{NN}^{wrap}()$ in \cite{torchspatial} on top of each location featurization method (i.e., WRAP, GeoCLIP, RFF, SatCLIP). The ResNet consists of \textbf{1)} a linear layer with ReLU activation, \textbf{2)} 4 residual blocks, each consisting of 2 linear layers with ReLU activation, followed by a skip connection. The output dimension of the ResNet is 256. For WRAP and RFF, the Fourier features are directly input to the ResNet. For GeoCLIP and SatCLIP, the features output from the pre-trained encoders are input to the ResNet and kept frozen. The total size of each location encoder are shown in Table \ref{tab:supp_loc_sizes}.    

\begin{table}[h]
\centering
\renewcommand{\arraystretch}{1.2}
\setlength{\tabcolsep}{4pt}
\footnotesize
\begin{tabular}{lccc}
\toprule
 & \textbf{Total Params} & \textbf{Trainable Params} & \textbf{ResNet Input Dim.}\\
\midrule
WRAP & 527.6 K & 527.6 K & 4\\
GeoCLIP & 10.11 M & 657.7 K & 512\\
RFF & 658.18 K & 657.7 K & 512\\
SatCLIP & 1.81 M & 592.1 K & 256\\
\bottomrule
\end{tabular}
\caption{Size of each location encoder.}
\label{tab:supp_loc_sizes}
\end{table}

\subsection{Fusion Methods}
In \cref{tab:supp_sizes_fmow,tab:supp_sizes_poverty,tab:supp_sizes_inat,tab:supp_sizes_yfcc} we report the total model size for each combination of dataset, fusion method, and location encoder. These results exclude the domain predictor sizes for each dataset which are reported in \cref{sec:supp_dp_size}. Note that the image-only base model used for ERM is the same as what is used for the IRM, CORAL, and GroupDRO results reported in \cref{tab:main_results}. Overall, these fusion methods do not incur a significant increase in model size compared to the image-only base model.

\begin{table}
\centering
\renewcommand{\arraystretch}{1.25}
\setlength{\tabcolsep}{5pt}
\footnotesize
\begin{tabular}{@{}p{1.6cm} l c c@{}}
\toprule
&
\textbf{Method} &
\textbf{Total Params} &
\textbf{Trainable Params}\\
\midrule
\multirow{2}{=}{\textbf{None}}
& ERM & 427.99 M & 427.99 M\\
& D\textsuperscript{3}G & 428.26 M & 428.26 M\\
\midrule
\multirow{4}{=}{\textbf{WRAP}}
& Concat & 428.54 M & 428.54 M \\
& Geo Priors & 428.54 M & 428.54 M \\
& FiLM & 429.31 M & 429.31 M \\
& D\textsuperscript{3}G & 428.86 M & 428.86 M \\
\midrule
\multirow{4}{=}{\textbf{GeoCLIP}}
& Concat & 438.11 M & 428.67 M\\
& Geo Priors & 438.11 M & 428.67 M \\
& FiLM & 438.89 M & 429.44 M\\
& D\textsuperscript{3}G & 438.44 M & 428.99 M \\

\bottomrule
\end{tabular}
\caption{Size of each model trained on WILDS-FMoW for different combinations of location encoder and fusion method. These sizes exclude the domain predictor.}
\label{tab:supp_sizes_fmow}
\end{table}

\begin{table}
\centering
\renewcommand{\arraystretch}{1.25}
\setlength{\tabcolsep}{5pt}
\footnotesize
\begin{tabular}{@{}p{1.6cm} l c c@{}}
\toprule
&
\textbf{Method} &
\textbf{Total Params} &
\textbf{Trainable Params}\\
\midrule
\multirow{2}{=}{\textbf{None}}
& ERM & 11.20 M & 11.20 M\\
& D\textsuperscript{3}G & 11.23 M & 11.23 M\\
\midrule
\multirow{4}{=}{\textbf{WRAP}}
& Concat & 11.73 M & 11.73 M \\
& Geo Priors & --- & --- \\
& FiLM & 12.12 M & 12.12 M \\
& D\textsuperscript{3}G & 11.83 M & 11.83 M \\
\midrule
\multirow{4}{=}{\textbf{GeoCLIP}}
& Concat & 21.31 M & 11.86 M\\
& Geo Priors & --- & --- \\
& FiLM & 21.70 M & 12.25 M\\
& D\textsuperscript{3}G & 21.40 M & 11.96 M \\
\bottomrule
\end{tabular}
\caption{Size of each model trained on WILDS-PovertyMap for different combinations of location encoder and fusion method. These sizes exclude the domain predictor.}
\label{tab:supp_sizes_poverty}
\end{table}

\begin{table}
\centering
\renewcommand{\arraystretch}{1.25}
\setlength{\tabcolsep}{5pt}
\footnotesize
\begin{tabular}{@{}p{1.6cm} l c c@{}}
\toprule
&
\textbf{Method} &
\textbf{Total Params} &
\textbf{Trainable Params}\\
\midrule
\multirow{2}{=}{\textbf{None}}
& ERM & 16.68 M & 16.68 M\\
& D\textsuperscript{3}G & 233.59 M & 233.59 M\\
\midrule
\multirow{4}{=}{\textbf{WRAP}}
& Concat & 19.29 M & 19.29 M \\
& Geo Priors & 2.62 M & 2.62 M \\
& FiLM & 21.94 M & 21.94 M \\
& D\textsuperscript{3}G & 234.19 M & 234.19 M \\
\midrule
\multirow{4}{=}{\textbf{GeoCLIP}}
& Concat & 28.87 M & 19.42 M\\
& Geo Priors & 12.20 M & 2.75 M \\
& FiLM & 31.51 M & 22.07 M\\
& D\textsuperscript{3}G & 243.77 M & 234.32 M \\
\bottomrule
\end{tabular}
\caption{Size of each model trained on iNat-Biomes for different combinations of location encoder and fusion method. These sizes exclude the domain predictor.}
\label{tab:supp_sizes_inat}
\end{table}

\begin{table}
\centering
\renewcommand{\arraystretch}{1.25}
\setlength{\tabcolsep}{5pt}
\footnotesize
\begin{tabular}{@{}p{1.6cm} l c c@{}}
\toprule
&
\textbf{Method} &
\textbf{Total Params} &
\textbf{Trainable Params}\\
\midrule
\multirow{2}{=}{\textbf{None}}
& ERM & 204.90 K & 204.90 K\\
& D\textsuperscript{3}G & 443.34 K & 443.34 K\\
\midrule
\multirow{4}{=}{\textbf{WRAP}}
& Concat & 758.12 K & 758.12 K \\
& Geo Priors & 553.32 K & 553.32 K \\
& FiLM & 5.46 M & 5.46 M \\
& D\textsuperscript{3}G & 1.04 M & 1.04 M \\
\midrule
\multirow{4}{=}{\textbf{GeoCLIP}}
& Concat & 10.34 M & 888.16 K\\
& Geo Priors & 10.13 M & 683.36 K \\
& FiLM & 15.04 M & 5.59 M\\
& D\textsuperscript{3}G & 10.62 M & 1.17 M \\

\bottomrule
\end{tabular}
\caption{Size of each model trained on TorchSpatial-YFCC for different combinations of location encoder and fusion method. These sizes exclude the domain predictor.}
\label{tab:supp_sizes_yfcc}
\end{table}

\subsection{Domain Predictors}
\label{sec:supp_dp_size}

Domain prediction applied to the location encoder is a lightweight addition to any image-location fusion method. For each dataset, we use a single linear layer for the domain predictor. The size of the domain predictor for different datasets is shown in Table \ref{tab:supp_dp_sizes}. Since the location encoder output dimension is always 256, these sizes are constant across all choices of location encoder.

\begin{table}[h]
\centering
\renewcommand{\arraystretch}{1.2}
\setlength{\tabcolsep}{6pt}
\footnotesize
\begin{tabular}{cccc}
\toprule
\textbf{FMoW} & \textbf{PovertyMap} & \textbf{iNat-Biomes} & \textbf{YFCC} \\
\midrule
1.5 K & 0.5 K & 3.6 K & 0.5 K \\
\bottomrule
\end{tabular}
\caption{Domain predictor model sizes (i.e., number of parameters) for each dataset.}
\label{tab:supp_dp_sizes}
\end{table}

%% file: sec/sup/dp_ablation_full.tex
\begin{table*}[t]
\centering
\renewcommand{\arraystretch}{1.25}
\setlength{\tabcolsep}{6pt}
\footnotesize
\begin{tabular}{@{}l l cc cc cc cc@{}}
\toprule
\multirow{2}{*}{\textbf{Method}} & 
\multirow{2}{*}{\textbf{Loc. Enc.}} &
\multicolumn{2}{c}{\textbf{FMoW}} &
\multicolumn{2}{c}{\textbf{PovertyMap}} &
\multicolumn{2}{c}{\textbf{iNat-Biomes}} &
\multicolumn{2}{c}{\textbf{YFCC}} \\
\cmidrule(lr){3-4}
\cmidrule(lr){5-6}
\cmidrule(lr){7-8}
\cmidrule(lr){9-10}
& &
\textbf{Avg} & \textbf{Worst} &
\textbf{Avg} & \textbf{Worst} &
\textbf{Avg} & \textbf{Worst} &
\textbf{Avg} & \textbf{Worst} \\
\midrule

\multirow{2}{*}{Concat} 
& WRAP    & 66.9 (0.3) & 48.1 (1.6) & 0.78 (0.02) & 0.48 (0.04) & 71.7 (0.1) & 68.4 (0.3) & 51.5 (0.0) & 48.3 (0.1) \\
& GeoCLIP & 66.0 (0.8) & 52.8 (1.0) & 0.80 (0.02) & 0.53 (0.03) & 74.6 (0.1) & 70.7 (0.6) & 52.4 (0.0) & 49.2 (0.1) \\[3pt]
\midrule

\multirow{2}{*}{Geo Priors}
& WRAP    & 65.7 (0.2) & 48.1 (1.7) & --- & --- & 72.5 (0.0) & 65.2 (0.2) & 50.5 (0.0) & 47.4 (0.0)\\
& GeoCLIP & 66.8 (0.1) & 49.6 (0.8) & --- & --- & 72.8 (0.0) & 64.7 (0.6) & 50.9 (0.0) & 47.9 (0.0) \\[3pt]
\midrule

\multirow{2}{*}{FiLM}
& WRAP    & 66.4 (0.2) & 49.4 (2.0) & 0.79 (0.02) & 0.50 (0.02) & 59.6 (0.7) & 52.3 (1.1) & 49.6 (0.1) & 46.2 (0.0) \\
& GeoCLIP & 65.52 (0.5) & 49.9 (1.7) & 0.82 (0.01) & 0.57 (0.02) & 48.3 (2.3) & 43.9 (3.7) & 49.6 (0.1) & 46.3 (0.1) \\[3pt]
\midrule

\multirow{2}{*}{D$^3$G}
& WRAP    & 66.4 (0.4) & 54.2 (1.6) & 0.78 (0.02) & 0.46 (0.02) & 66.1 (0.8) & 46.8 (8.4) & 51.4 (0.0) & 48.3 (0.1) \\
& GeoCLIP & 70.0 (0.8) & 54.0 (0.9) & 0.78 (0.02) & 0.45 (0.02) & 61.9 (6.3) & 53.6 (10.1) & 51.4 (0.0) & 48.2 (0.0)\\
\bottomrule
\end{tabular}
\caption{Ablation results for the auxiliary domain prediction loss. The table shows both overall average and worst-group performance results when no domain prediction loss is applied (i.e. $\alpha = 0$) across different combinations of fusion method and location encoder.}
\label{tab:dp_ablation_full}
\end{table*}

%% file: sec/sup/loc_enc_ablation_full.tex
\begin{table*}
\centering
\renewcommand{\arraystretch}{1.25}
\setlength{\tabcolsep}{6pt}
\footnotesize
\begin{tabular}{@{}p{1.4cm} l cc cc cc cc@{}}
\toprule
\multirow{2}{*}{\textbf{Method}} &
\multirow{2}{*}{\textbf{Loc. Enc.}} &
\multicolumn{2}{c}{\textbf{FMoW}} &
\multicolumn{2}{c}{\textbf{PovertyMap}} &
\multicolumn{2}{c}{\textbf{iNat-Biomes}} &
\multicolumn{2}{c}{\textbf{YFCC}} \\
\cmidrule(lr){3-4}
\cmidrule(lr){5-6}
\cmidrule(lr){7-8}
\cmidrule(lr){9-10}
& &
\textbf{Avg} & \textbf{Worst} &
\textbf{Avg} & \textbf{Worst} &
\textbf{Avg} & \textbf{Worst} &
\textbf{Avg} & \textbf{Worst} \\
\midrule

\multirow{3}{=}{Concat}
& None      & 66.2 (0.2) & 50.9 (0.1) & 0.78 (0.02) & 0.40 (0.02) & 57.4 (0.2) & 49.9 (4.7) & 51.5 (0.0) & 48.3 (0.0) \\
& WRAP      & 66.8 (0.4) & 49.1 (1.6) & 0.80 (0.02) & 0.52 (0.04) & 73.1 (0.0) & 70.2 (0.2) & 51.6 (0.1) & 48.4 (0.2) \\
& GeoCLIP   & 66.2 (0.2) & 52.0 (0.3) & 0.80 (0.02) & 0.54 (0.02) & 73.6 (0.0) & 71.1 (0.1) & 52.4 (0.0) & 49.2 (0.0) \\
\midrule

\multirow{3}{=}{Geo Priors}
& None      & 65.5 (0.6) & 50.0 (1.4) & --- & ---  & 59.1 (2.5) & 43.7 (8.83) & 50.2 (0.0)  & 47.1 (0.1) \\
& WRAP      & 65.7 (0.2) & 49.1 (2.0) & --- & --- & 72.6 (0.03) & 65.5 (0.2) & 50.5 (0.0) & 47.4 (0.1) \\
& GeoCLIP   & 66.8 (0.1) & 50.9 (1.1) & --- & --- & 72.6 (0.03) & 66.9 (0.5) & 50.9 (0.0) & 47.9 (0.0) \\
\midrule

\multirow{3}{=}{FiLM}
& None      & 64.5 (0.2) & 50.0 (2.0) & 0.80 (0.02) & 0.43 (0.03) & 56.8 (1.0)  & 52.1 (1.4) & 50.2 (0.1) & 46.9 (0.1) \\
& WRAP      & 66.9 (0.5) & 51.8 (2.1) & 0.80 (0.02) & 0.52 (0.04) & 61.5 (1.2) & 56.0 (1.1) & 49.7 (0.0) & 46.3 (0.1) \\
& GeoCLIP   & 66.9 (0.8) & 51.7 (2.0) & 0.82 (0.02) & 0.57 (0.03) & 53.1 (1.5) & 49.8 (1.6) & 49.7 (0.2) & 46.2 (0.2) \\
\midrule

\multirow{3}{=}{D\textsuperscript{3}G}
& None      & 66.7 (0.2)  & 51.7 (1.2) & 0.79 (0.02) & 0.46 (0.04) & 63.2 (1.4) & 44.9 (8.2) & 51.4 (0.0) & 48.3 (0.1) \\
& WRAP      & 66.9 (0.2) & 53.1 (0.7) & 0.79 (0.02) & 0.47 (0.04) & 65.2 (1.6) & 47.0 (1.4) & 51.3 (0.1) & 48.2 (0.2) \\
& GeoCLIP   & 66.6 (0.1) & 55.8 (0.7) & 0.79 (0.02) & 0.46 (0.04) & 60.7 (2.7) & 55.1 (0.7) & 51.3 (0.0) & 48.2 (0.1) \\
\bottomrule
\end{tabular}
\caption{Ablation results for the choice of location encoder, showing both overall average and worst-group performance.}
\label{tab:loc_enc_ablation_full}
\end{table*}

%% file: sec/sup/additional_loc_encoders_full.tex
\begin{table*}[t]
\centering
\renewcommand{\arraystretch}{1.25}
\setlength{\tabcolsep}{10pt}
\footnotesize
\begin{tabular}{@{}p{1.5cm} l l cc cc@{}}
\toprule
&
\multirow{2}{*}{\textbf{Loc. Enc.}} &
\multirow{2}{*}{\textbf{Method}} &
\multicolumn{2}{c}{\textbf{PovertyMap}}&
\multicolumn{2}{c}{\textbf{iNat-Biomes}}\\
\cmidrule(lr){4-5}
\cmidrule(lr){6-7}
& & &
\textbf{Avg} & \textbf{Worst} &
\textbf{Avg} & \textbf{Worst} \\
\midrule

\multirow{4}{=}{\textbf{Baselines}}
& RFF      & Concat      & 0.80 (0.02) & 0.53 (0.05) & \underline{74.4 (0.1)} & \underline{70.5 (0.2)} \\
& RFF      & Geo Priors  & ---  & ---         & 72.6 (0.0) & 65.7 (0.8) \\
& SatCLIP  & Concat      & 0.79 (0.01) & 0.48 (0.03) & 72.6 (0.1) & 68.6 (0.5) \\
& SatCLIP  & Geo Priors  & --- & ---         & 71.4 (0.1) & 64.0 (0.2) \\
\midrule

\multirow{8}{=}{\textbf{Ours\\(all w/ DP)}}
& RFF      & Concat      & 0.79 (0.02) & 0.54 (0.03) & \textbf{75.1 (0.0)} & \textbf{71.8 (0.2)} \\
& RFF      & Geo Priors  & --- & ---         & 72.7 (0.0) & 65.5 (0.6) \\
& RFF      & FiLM        & \underline{0.81 (0.02)} & \textbf{0.56 (0.03)} & 61.2 (1.8)  & 55.8 (0.6) \\
& RFF      & D\textsuperscript{3}G & 0.79 (0.02) & 0.46 (0.04) & 66.8 (0.7) & 54.7 (7.0) \\
& SatCLIP  & Concat      & 0.80 (0.02)  & 0.50 (0.03) & 72.6 (0.0) & 69.0 (0.4) \\
& SatCLIP  & Geo Priors  & --- & ---         & 71.5 (0.1) & 64.2 (0.4) \\
& SatCLIP  & FiLM        & \textbf{0.82 (0.02)} & \underline{0.55 (0.05)} & 69.5 (0.1) & 64.3 (0.6) \\
& SatCLIP  & D\textsuperscript{3}G & 0.79 (0.02) & 0.48 (0.04) & 65.9 (0.7) & 62.6 (0.5) \\
\bottomrule
\end{tabular}
\caption{Full results on PovertyMap and iNat-Biomes with RFF and SatCLIP location encoders.}
\label{tab:additional_loc_enc_full}
\end{table*}